\newcommand{\ig}[1]{}

\documentclass[12pt,letterpaper]{article}

\usepackage{graphicx}
\usepackage[hyphens]{url}
\usepackage{framed}
\usepackage{rotating}
\usepackage{named}
\usepackage{multirow}

\begin{document}

\title{Weight-Based Variable Ordering in the Context of High-Level
  Consistencies}

\author{Robert~J.\ Woodward
\and Berthe~Y.~Choueiry\\
\\
Constraint Systems Laboratory\\ University of Nebraska-Lincoln, USA\\
{\tt \{rwoodwar$\mid$choueiry\}@cse.unl.edu}\\
}

\date{\today}

\maketitle

\begin{abstract}
Dom/wdeg is one of the best performing heuristics for dynamic variable
ordering in backtrack search \cite{boussemart:ecai04}.  As originally
defined, this heuristic increments the weight of the constraint that
causes a domain wipeout (i.e., a dead-end) when enforcing arc
consistency during search. ``The process of weighting constraints with
dom/wdeg is not defined when more than one constraint lead to a domain
wipeout \cite{Vion2011}.'' In this paper, we investigate how weights
should be updated in the context of two high-level consistencies,
namely, singleton (POAC) and relational consistencies (RNIC).  We
propose, analyze, and empirically evaluate several strategies for
updating the weights.  We statistically compare the proposed
strategies and conclude with our recommendations.
\end{abstract}

\section{Introduction}

Variable-ordering heuristics are critical for the effectiveness of
backtrack search to solve Constraint Satisfaction Problems (CSPs).
Common heuristics implement the fail-first principal, choosing the
most constrained variable as the next variable to assign.  One such
heuristic is dom/ddeg, which selects the variable with the smallest
ratio of its current domain to its future degree.  A more recent
heuristic, dom/wdeg, uses the weighted degree of a variable by
assigning a weight, initially set to one, to each constraint, and
incrementing this weight whenever the constraint causes a domain
wipeout~\cite{boussemart:ecai04}.  Recently, higher-level
consistencies (HLC) have shown promise as lookahead for solving
difficult CSPs~\cite{bennaceurPOAC2001,Robert:AAAI11,Robert:cp12,amineAAAI14}.

Because HLC algorithms typically consider more than one constraint at
the same time, updating the weights of the constraints in dom/wdeg is
currently an open question \cite{Vion2011}.  This paper focuses on
answering this question in the context of two high-level
consistencies, namely, Partition-One Arc-Consistency (POAC)
\cite{bennaceurPOAC2001} and Relational Neighborhood Inverse
Consistency (RNIC) \cite{Robert:AAAI11}.  Our study focuses on these
two consistencies because they have both been shown to be beneficial
when used for lookahead during search.

For POAC and RNIC we introduce four and three strategies,
respectively, to increment the weights of the constraints.  For both
consistencies we find that a baseline strategy corresponding to the
original dom/wdeg proposal is statistically the worst of the proposed
strategies.  We conclude the high-level consistency should influence
the weights.  For POAC we find that the proposed strategy {\sc AllS} is
statistically the best.  For RNIC the two non-baseline strategies are
statistically equivalent.

Other popular variable-ordering heuristics include Impact-Based Search
\cite{Refalo04:Impact} and Activity-Based Search
\cite{Michel2012:activity}.  These heuristics rely on information
about the domain filtering resulting from enforcing a given consistency.
Because they ignore the operations of the consistency algorithm, it is
not clear how these heuristics could be used to order the propagation
queue of the consistency
algorithm~\cite{Wallace92orderingheuristics,amineAAAI14}.  Further,
it is also not clear how to apply them in the context of consistency
algorithms that filter the relations~\cite{Robert:AAAI11,Robert:cp12}.

The paper is structured as follows.  Section~\ref{sec:bg} summarizes
relevant background information. Section~\ref{sec:weight} introduces
our weighting schemes for POAC and RNIC and
Section~\ref{sec:experiments} empirically evaluates them.  Finally,
Section~\ref{sec:conclusion} concludes the paper.

\section{Background}
\label{sec:bg}

A Constraint Satisfaction Problem (CSP) is defined by
$\mathcal{P} = (\mathcal{X},\mathcal{D},\mathcal{C})$.  $\mathcal{X}$
is a set of variables where a variable $x_i \in \mathcal{X}$ has a
finite domain $dom(x_i) \in \mathcal{D}$.  A constraint
$c_i \in \mathcal{C}$ is specified by its scope $scp(c_i)$ and its
relation $rel(c_i)$.  $scp(c_i)$ is the set of variables to which
$c_i$ applies and $rel(c_i)$ is the set of allowed tuples.  A tuple on
$scp(c_i)$ is consistent with $c_i$ if it belongs to
$rel(c_i)\cap \prod_{x_i \in scp (c_i)} dom(x_i)$.  A solution to the
CSP assigns, to each variable, a value taken from its domain such that
all the constraints are satisfied.  The problem is to determine the
existence of a solution and is known to be NP-complete.  To this day,
backtrack search remains the only known sound and complete algorithm
for solving CSPs \cite{Bitner:BT:1975}.  Search operates by assigning
a value to a variable and backtracks when a dead-end is
encountered\ig{ by undoing past assignments}.  The variable-ordering
heuristic determines the order that variables are assigned in search,
which can be dynamic (i.e., change during search).
\citeauthor{boussemart:ecai04}\ \shortcite{boussemart:ecai04}
introduced dom/wdeg, a popular dynamic variable-ordering heuristic.
This heuristic associates to each constraint $c \in \mathcal{C}$ a
weight $w_c(c)$, initialized to one, that is incremented by one
whenever the constraint causes a domain wipeout when enforcing arc
consistency.  The next variable $x_i$ chosen by dom/wdeg is the one
with the smallest ratio of current domain size to the weighted degree,
$\alpha_{wdeg}(x_i)$, given by
  \begin{equation}
\alpha_{wdeg}(x_i) = \sum_{(c \in \mathcal{C}_f) \land (x_i \in \mathit{scp}(C))} w_c(c)
\label{eq:alpha-wdeg}
  \end{equation}
where $\mathcal{C}_f \subseteq \mathcal{C}$ is the set of constraints
with at least two future variables (i.e., variables who have not been
assigned by search).

Modern solvers
enforce a given consistency property on the CSP
after each
variable assignment.
This lookahead removes from the
domains of the unassigned variables values that cannot participate in
a solution.  Such filtering prunes from the search space fruitless
subtrees, reducing thrashing and the size of the search space.
The higher the consistency level enforced during
lookahead, the stronger the pruning and the smaller the search space.

The
standard property for lookahead is Generalized Arc
Consistency (GAC)~\cite{Mackworth:77IJ}.  A CSP is GAC iff, for every
constraint $c_i$, and every variable $x \in scp(c_i)$, every value $v
\in dom(x)$ is consistent with $c_i$ (i.e., appears in some consistent
tuple of $rel_i$).
Singleton Arc-Consistency (SAC) ensures that
no  domain becomes empty when  enforcing GAC
after assigning a value to a
variable
\cite{Debruyne:97ijcai:sac}. This operation is called a {\em
  singleton test\/}.  Algorithms for enforcing SAC remove all domain
values that fail the singleton test.  Partition-One Arc-Consistency
(POAC) adds an additional condition to SAC \cite{bennaceurPOAC2001}.
Let $(x_i, v_i)$ denotes a variable-value pair, $(x_i, v_i) \in
\mathcal{P}$ iff $v_i \in dom(x_i)$.  A constraint network
$\mathcal{P}=(\mathcal{X,D,C})$ is Partition-One Arc-Consistent (POAC)
iff $\mathcal{P}$ is SAC and for all $x_i \in \mathcal{X}$, for all
$v_i \in dom(x_i)$, for all $x_j \in X$, there exists $v_j \in
dom(x_j)$ such that $(x_i,v_i) \in \mathrm{GAC}(\mathcal{P} \cup \{x_j
\gets v_j\})$, where $\mathrm{GAC}(\mathcal{P} \cup \{x_j \gets
v_j\})$ is the CSP after assigning $x_j \gets v_j$ and running
GAC~\cite{bennaceurPOAC2001}.

Using the terminology of
\citeauthor{Debruyne:97ijcai:sac}\ \shortcite{Debruyne:97ijcai:sac},
we say that a consistency property $p$ is \emph{stronger} than $p'$ if
in any CSP where $p$ holds $p'$ also holds.  Further, we say that $p$
is \emph{strictly stronger} than $p'$ if $p$ is stronger than $p'$,
and there exists at least one CSP in which $p'$ holds but $p$ does
not.  We say that $p$ and $p'$ are equivalent if $p$ is stronger than
$p'$, and vice versa.  Finally, we say that $p$ and $p'$ are
incomparable when there exists at least one CSP in which $p$ holds but
$p'$ does not, and vice versa.  In practice, when a consistency
property $p$ is stronger than another $p'$, enforcing $p$ never yields
less pruning than enforcing $p'$ on the same problem.
POAC is strictly stronger than SAC and
SAC than GAC.

\citeauthor{amineAAAI14}\ \shortcite{amineAAAI14} introduced two
algorithms for enforcing POAC: POAC-1 and its adaptive version APOAC.
POAC-1 operates by enforcing SAC.  When running a singleton test on
each of the values in the domain of a given variable, POAC-1 maintains
a counter for each value in the domain of the remaining variables to
determine whether or not the corresponding value was removed by any of
the singleton tests.  Values that are removed by each of those
singleton tests are identified as not POAC and removed from their
respective domains.  POAC-1 was found to reach quiescence faster than
SAC.  In POAC-1, all the CSP variables are singleton tested and the
process is repeated over all the variables until a fixpoint is
reached.  In APOAC, the adaptive version of POAC-1, the process is
interrupted as soon as a given number of variables are singleton
tested.  This number depends on input parameters and is updated by
learning during search.

Neighborhood Inverse Consistency (NIC)~\cite{freuder96aaai} ensures
that every value in the domain of a variable $x_i$ can be extended to a
solution of the subproblem induced by $x_i$ and the variables
in its neighborhood.  In the dual graph of a CSP, the vertices represent the CSP
constraints and the edges connect vertices representing constraints
whose scopes overlap.  Relational Neighborhood Inverse Consistency
(RNIC)~\cite{Robert:AAAI11} enforces NIC on the dual graph of the CSP.
That is, it ensures that any tuple in any relation can be extended in
a consistent assignment to all the relations in its neighborhood in
the dual graph.  NIC and RNIC are theoretically incomparable
\cite{Robert:cp12}, but RNIC has two main advantages over NIC.  First,
NIC was originally proposed for binary CSPs and the neighborhoods in
NIC likely grow too large on non-binary CSPs; second, RNIC can
operate on different dual graph structures to save time.  Three
variations of RNIC were introduced, wRNIC, triRNIC, and wtriRNIC,
which operate on modified dual graphs.  Given an instance, selRNIC
uses a
decision tree to automatically select the dual graph for RNIC to operate on.

\section{Weighting Schemes}
\label{sec:weight}

We introduce weighting schemes first in the context of singleton
consistencies, namely Partition-One Arc-Consistency (POAC), and then
in that of relational consistencies, namely Relational Neighborhood
Inverse Consistency (RNIC).

Enforcing a high-level consistency (HLC) property is typically
costlier than enforcing GAC, but typically yields more powerful
pruning.  Further, it is often more effective, in terms of CPU time, to run a GAC
before an HLC algorithm \cite{Debruyne:97ijcai:sac}, as we choose
to do in this paper.

\subsection{Partition-One Arc-Consistency}
\label{sec:singletonweight}

We first investigate the case of POAC, which operates by initially running
a GAC algorithm then applying the following operation to each variable
until no change occurs.  For a given variable, it applies a singleton
test to each value in the domain of the variable.  A singleton test
assigns the value to the variable and enforces GAC on the problem.  We
propose four strategies to increment weights during POAC:
\begin{description}
\item[{\sc Old}:] We allow only the GAC call before POAC to increment
  the weight of the constraint that causes a domain wipeout.  That is,
  POAC is not allowed to alter the weights.  This strategy is the
  simplest and it is a direct application of the original proposal
  \cite{boussemart:ecai04}.  In our experiments we use this strategy
  as a baseline and show it does not perform well in practice.

\item[{\sc AllS}:] In addition to incrementing the weights according
  the above strategy (i.e., {\sc Old}), we allow every singleton test
  to increment the weight of a constraint whenever enforcing GAC on
  this constraint during the singleton test directly wipes out the
  domain of a variable.  This update is made at most once for each
  singleton test.  Under this strategy, all constraints that caused
  domain wipeouts are affected, thus, we call it {\sc AllS}.  Notice
  that the weight of more than one constraint may be updated even
  though search does not have to backtrack.  This behavior differs
  from the original proposal \cite{boussemart:ecai04}.

\item[{\sc LastS}:] In addition to incrementing the weights according
  to {\sc Old}, we increment the weight of the constraint causing a
  domain wipeout at the {\em last\/} singleton test on a given
  variable if and only if all previous singleton tests on the values
  of this variable have failed.  Thus, we only increment the weight of
  a single constraint and do so only when search has to backtrack,
  which conforms to the spirit of the original heuristic.  Notice, the
  order of values singleton tested affects this strategy.

\item[{\sc Var}:] This strategy encapsulates {\sc Old} as a first
  step and increments the weight of the \emph{variable} on which all
  singleton tests have failed (thus forcing search to backtrack).  In
  order to implement this strategy we add a counter for the weight of
  each variable $\mathit{w_v}$, initially zero.  When a variable fails
  all of its singleton tests during propagation the counter $\mathit{w_v}$ for
  that variable is incremented by one.  We propose to integrate
  $\mathit{w_v}$ with the weighted degree function of dom/wdeg as
  follows:
  \begin{equation}
\alpha_{wdeg}^{\normalfont\textsc{\scriptsize Var}}(x_i) = w_v(x_i) + \sum_{(c \in \mathcal{C}_f) \land (x_i \in \mathit{scp}(c))} w_c(c)
\label{eq:alpha-var}
  \end{equation}
\ig{\[\alpha_\mathrm{\sc Var}(x_i) = w_v[x_i] + \sum_{
  \begin{array}{c}
    \scriptstyle c \in \mathcal{c}_f \land\\
    \scriptstyle x_i \in \mathit{scp}(c)
   \end{array}} w_c[c]\]}
  where $\mathcal{C}_f \subseteq \mathcal{C}$ is the set of
  constraints with at least two future variables.  The rationale
  behind this strategy is the following.  The goal of the heuristic
  dom/wdeg is to identify the conflicts in the problem and address
  them earlier, rather than later, in the search.  {\sc Var} puts the
  blame on the variable that first caused the failure of POAC.
\end{description}

\subsection{Relational Neighborhood Inverse Consistency}
\label{sec:w-rnic}
The relational consistency property RNIC is equivalent to enforcing
Neighborhood Inverse Consistency (NIC) on the dual graph of the CSP
\cite{freuder96aaai,Robert:AAAI11}.  The RNIC property ensures that
every tuple in every relation can be extended to a solution in the
subproblem induced on the dual graph of the CSP by the relation and
its neighboring relations.  The RNIC algorithm operates on table
constraints and removes, from a given relation, all the tuples that do
not appear in a solution in the induced (dual) CSP of its neighborhood
\cite{Robert:AAAI11}.  We propose three strategies to increment
weights when RNIC is used for lookahead during
search:
\begin{description}
\item[{\sc Old}:] As in POAC in Section~\ref{sec:singletonweight}, we
  allow only the GAC call (preceding the call to RNIC) to increment the
  weight of the constraint that causes domain wipeout.

\item[{\sc AllC}:] This strategy encapsulates {\sc Old} as a first
  step.  During lookahead, RNIC is called on each constraint with two
  or more future variables.  When the RNIC algorithm removes all the
  tuples of a given relation, {\sc AllC} increments the weights of all
  the relations in the induced (dual) CSP.  The rationale being that
  this considered combination of relations (which is the relation and
  its neighborhood in the dual graph) is `collectively' responsible
  for the `relation' wipeout.

\item[{\sc Head}:] This strategy is similar to {\sc AllC}, except that
  we increment only the weight of the constraint whose relation was
  emptied by the RNIC algorithm and do not increment the weights of
  its neighborhood in the dual graph.
\end{description}

\section{Experimental Evaluation}
\label{sec:experiments}

We evaluate the effectiveness of the strategies proposed for POAC and
RNIC in Sections~\ref{sec:exp-poac} and~\ref{sec:exp-rnic},
respectively.

\subsection{Experimental Setup}
\label{sec:setup}
We consider the problem of finding a single solution to a CSP using
backtrack search with some lookahead, $d$-way branching, dom/wdeg
dynamic variable-ordering heuristic \cite{boussemart:ecai04}, and
lexicographic value ordering.  We use STR2+ for enforcing GAC
\cite{lecoutreSTR2}, APOAC for enforcing POAC
\cite{amineAAAI14},\footnote{Using the terminology of Balafrej et
  al.~\cite{amineAAAI14}, we use the following parameters and their
  recommended values for APOAC $maxK = n$, last drop with
  $\beta = 0.05$, and 70\%-PER. Where $maxK$ indicates the number of
  processed items in the propagation queue, $\beta$ is the threshold
  of search-space reduction during the learning phase and 70\%-PER is
  the percentile for learning the value of $maxK$.}  and selRNIC for
enforcing RNIC \cite{Robert:AAAI11}.  We use the benchmark problems
available from Lecoutre's
website.\footnote{\url{www.cril.univ-artois.fr/~lecoutre/benchmarks.html}}
Benchmarks are selected separately for POAC and RNIC.  For a given
consistency level, if any instance is solved by any of the weighing
schemas of the considered consistency within the time limit of 60
minutes and memory limit of 8GB, then the entire benchmark is included
in the experiment.  For benchmarks in intension we convert the
instance to extension prior to solving and do not include the time for
conversion.\footnote{In a study not reported we found that STR2+ is
  faster at solving CSP instances than running GAC on the original
  intension constraints because STR explores the satisfying tuples
  instead of valid tuples.  As STR and RNIC algorithms require table
  constraints we pre-convert the instances.  The conversion time is
  the same for each algorithm and can safely be ignored.}  From the
254 benchmark problems (total 8,549 instances) available on Lecoutre's
website, our results are reported on 144 benchmarks (total 4,233
instances) for POAC and 132 (total 3,869 instances) for RNIC.

We summarize the results of these experiments in
Tables~\ref{table:all-poac}--\ref{table:rnic-gc} and
Figures~\ref{fig:poac-time} and~\ref{fig:rnic-time}.  For each
strategy, we report in
Tables~\ref{table:all-poac}--\ref{table:rnic-gc}:
\begin{itemize}
\item
The number of completions (\# Completions) with the total number of
instances in parenthesis.

\item
The sum of the CPU time in seconds ($\sum$CPU sec.) computed over
instances where at least one algorithm terminated (given in
parenthesis).  When an algorithm does not terminate within 60 minutes,
we add 3,600 seconds to the CPU time and indicate with a $>$ sign that
the time reported is a lower bound.  We boldface the smallest CPU
time.

\item
The average number of node visits (Average NV) computed over the
instances where all strategies completed (given in parenthesis).
\end{itemize}
Figures~\ref{fig:poac-time} and~\ref{fig:rnic-time} plot the number of
instances solved by each strategy (Y-axis) as the CPU time increases
(X-axis).

In addition to the above experiment, we also conduct a statistical
analysis of the relative performance of the proposed strategies.  We
compare pairwise the strategies corresponding to each higher-level
consistency (i.e., POAC and RNIC) in order to determine whether or not
a statistical difference exists between the strategies.  Because
search may fail to complete within the time limit, we consider our
results to be right-censored and analyze them using a nonparameterized
Wilcoxon signed-rank test \cite{Wilcoxon:1945}.  The test operates by
comparing the rank of the differences of the paired data.  Differences
of zero have no effect on the test and are safely discarded before
ranking.  Further, given the clock precision, we discard data points
where the CPU difference is less than one second.  We assume a
one-tailed distribution and significance level of
$p=0.05$.\footnote{Check Palmieri et al.~\cite{Palmieri2016} for an
  overview of the Wilcoxon signed-rank test and the adopted
  methodology.}  In the presence of censored data, we adopt the
following procedure to generate the data for each pairwise test.
First, we run each strategy on each instance for the time limit (i.e.,
60 minutes).  If both strategies solve the instance, the data is
included in the analysis.  If neither strategy solves the instance,
the instance is excluded from the analysis (i.e., the difference is
zero and discarded).  If one strategy completes within the time
threshold and the other does not, we re-run the second strategy with
double the time limit (i.e., 120 minutes), recording this limit as the
completion time in case search does not terminate earlier.  By
allowing the additional time, the censored data no longer affects the
significance of the analysis \cite{Palmieri2016}.\footnote{Our
  approach is similar to that of Palmieri et al.~\cite{Palmieri2016}
  except that we exclude instances that neither strategy completes
  with the original time limit.}  The results obtained with the
doubled time limit are used only for the statistical analysis ranking
the relative performance of the strategies
(Table~\ref{tab:ranking-poac} and Expression~(\ref{eq:rnic-ranking})),
but not used for the results reported in
Tables~\ref{table:all-poac}--\ref{table:rnic-gc}.

\subsection{Partition-One Arc-Consistency}
\label{sec:exp-poac}

Based on the statistical analysis comparing the relative
performance for {\sc Old}, {\sc AllS}, {\sc LastS}, and {\sc Var} for
POAC, we conclude that {\em overall} (Table~\ref{tab:ranking-poac}):
\begin{itemize}
\item {\sc AllS} outperforms all others strategies
\item
{\sc LastS} and {\sc Var} are equivalent
\item
{\sc Old} exhibits the worst performance of the four strategies,
showing that it is important for dom/wdeg to increment the weights
with POAC, which justifies our investigations.
\end{itemize}
{ \def\arraystretch{1.5}
\begin{table}
  \caption{\small Statistical analysis of weighting schemes for POAC}
  \centerline{
      \small
  \begin{tabular}{|l|ccccccc|}\hline
    \multicolumn{1}{|l|}{\bf Benchmark} & \multicolumn{7}{c|}{\bf Ranking}\\ \hline \hline
    All benchmarks, put together & {\sc AllS} &$>$& {\sc LastS}& $\equiv$& {\sc Var}& $>$ &{\sc Old}\\ \hline
    `QCP/QWH,'  `BQWH' & \multirow{2}{*}{\sc LastS}& \multirow{2}{*}{$>$}& \multirow{2}{*}{\sc Old}& \multirow{2}{*}{$>$}& \multirow{2}{*}{\sc AllS} &\multirow{2}{*}{$\equiv$} &\multirow{2}{*}{\sc Var}\\
    ~~~(quasi-group completion) & & & & & & & \\
    `Graph Coloring' & {\sc Var} &$>$& {\sc AllS}& $>$ &{\sc LastS}& $>$& {\sc Old}\\
    `RAND' (random)  & {\sc Var}& $>$& {\sc AllS}& $\equiv$& {\sc LastS} &$\equiv$ &{\sc Old}\\
    `Crossword'  & {\sc Var}& $>$& {\sc AllS}& $\equiv$& {\sc LastS} &$\equiv$ &{\sc Old}\\\hline
  \end{tabular}
  }
\label{tab:ranking-poac}
\end{table}
}
However, a careful study of the individual benchmarks shows that {\sc
  LastS} on many quasi-group completion benchmarks and {\sc Var} are
competitive on many, but {\em not\/} all, graph coloring, random, and
crossword benchmarks.\footnote{Using the categories identified on
  Lecoutre's website.}  Re-running the statistical analysis on each
group of those benchmarks yields the results shown in the last four
rows of Table~\ref{tab:ranking-poac}.  Again, we insist that even
when considering individual benchmarks, the performance of {\sc AllS}
remains {\em globally\/} the most robust and consistent of all four
strategies.

Table~\ref{table:all-poac} summarizes the experiments' results on the
144 tested benchmarks.  { \def\arraystretch{1.5}
\begin{table}[!ht]
  \caption{\small Overall results of experiments for POAC}
  \label{table:all-poac}
  \centerline{
    \begin{tabular}{|l|rrrr|}\hline
      & \multicolumn{1}{c}{\sc Old} & \multicolumn{1}{c}{\sc AllS} & \multicolumn{1}{c}{\sc LastS} & \multicolumn{1}{c|}{\sc Var} \\
\hline
\hline
Completion (4,233) & 2,804 & 2,822 & 2,814 & 2,811 \\
$\sum$CPU sec. (2,846) &  $>$1,139,552  &  \textbf{$>$1,033,699}  &  $>$1,075,640  &  $>$1,065,547  \\
Average NV (2,775) & 19,181 & 16,712 & 16,503 & 21,875 \\\hline
    \end{tabular}
  }
\end{table}}
In terms of the number of completed instances and the CPU time, {\sc
AllS} is the best (with 2,822 instances and $>$1,033,699 seconds)
and {\sc Old} is the worst (with 2,804 instances and $>$1,139,552
seconds) of the four proposed strategies.  In terms of the average
number of nodes visited (i.e., reduction of the search space), {\sc
  LastS} visits the least amount of nodes on average (16,503),
followed by {\sc AllS} (16,712), {\sc Old} (19,181), and {\sc Var}
(21,875).\footnote{We offer the following hypothesis as to why {\sc
    Var} has the largest average of nodes visited.  The heuristic
  dom/wdeg is a `conflict-directed' heuristic in that it attempts to
  select the variable that participates in the largest number of
  `wipeouts.'  By incrementing the weight of the variable being
  singleton-tested, {\sc Var} perhaps increases the importance of a
  variable that `sees' the conflict rather than those variables that
  `cause' the conflict.  This hypothesis deserves a more thorough
  investigation.}

Table~\ref{table:poac-quasi} summarizes individual benchmark results
for the quasi-group completion category.  Compared to the quasi-group
completion analysis in Table~\ref{tab:ranking-poac}, the benchmarks
typically follow the statistical trend with {\sc LastS} performing the
best on the QCP-15 and QWH-20 benchmarks.  However, although {\sc
  LastS} was statistically the best, on bqwh-15-106, {\sc AllS} was
the fastest.
\begin{table}[!ht]
  \caption{\small Examples of quasi-group completion benchmark for POAC}
  \label{table:poac-quasi}
  \centerline{
    \begin{tabular}{|ll|rrrr|}\hline
{\bf Benchmark} & \multicolumn{1}{l|}{} & \multicolumn{1}{c}{\sc Old} & \multicolumn{1}{c}{\sc AllS} & \multicolumn{1}{c}{\sc LastS} & \multicolumn{1}{c|}{\sc Var} \\
\hline
\hline
\multicolumn{6}{|c|}{\textit{Where {\sc LastS} performs best}} \\
\multirow{3}{*}{QCP-15} & Completion (15) & 15    & 15    & 15    & 15 \\
      & $\sum$CPU sec. (15) &  3,920  &  5,480  &  \textbf{3,214}  &  6,083  \\
      & Average NV (15) & 30,488 & 38,641 & 23,963 & 33,589 \\
\hline
\multirow{3}{*}{QWH-20} & Completion (10) & 9     & 9     & 9     & 9 \\
      & $\sum$CPU sec. (9) &  6,625  &  7,329  &  \textbf{5,631}  &  12,337  \\
      & Average NV (9) & 57,453 & 58,623 & 45,095 & 63,225 \\
\hline\hline
\multicolumn{6}{|c|}{\textit{$\ldots$ but {\sc AllS} can still win on such benchmarks}} \\
\multirow{3}{*}{bqwh-15-106} & Completion (100) & 100   & 100   & 100   & 100 \\
      & $\sum$CPU sec. (100) &  196  &  \textbf{167}  &  189  &  211  \\
          & Average NV (100) & 599   & 433   & 531   & 507 \\
\hline
\end{tabular}
  }
\end{table}

Table~\ref{table:poac-gc-rand} summarizes individual benchmarks for
graph coloring, random, and crossword benchmarks.  {
\begin{table}[!ht]
  \caption{\small Examples of graph coloring, random, crossword benchmarks for POAC}
  \label{table:poac-gc-rand}
  \centerline{
    \begin{tabular}{|ll|rrrr|}\hline
{\bf Benchmark} & \multicolumn{1}{c|}{} & \multicolumn{1}{c}{\sc Old} & \multicolumn{1}{c}{\sc AllS} & \multicolumn{1}{c}{\sc LastS} & \multicolumn{1}{c|}{\sc Var} \\
\hline
\hline
\multicolumn{6}{|c|}{\textit{Where {\sc Var} performs best}} \\
\multirow{3}{*}{full-insertion} & Completion (41) & 28    & 28    & 28    & 29 \\
      & $\sum$CPU sec. (29) &  $>$12,720  &  $>$10,055  &  $>$10,182  &  \textbf{7,238}  \\
      & Average NV (28) & 16,725 & 12,676 & 13,312 & 8,749 \\
\hline
\multirow{3}{*}{tightness0.8} & Completion (100) & 98    & 97    & 97    & 99 \\
      & $\sum$CPU sec. (99) &  $>$59,907  &  $>$53,042  &  $>$56,945  &  \textbf{41,848}  \\
      & Average NV (97) & 1,213 & 1,085 & 1,196 & 1,315 \\
\hline
\multirow{3}{*}{wordsVg} & Completion (65) & 55    & 56    & 54    & 59 \\
      & $\sum$CPU sec. (59) &  $>$24,376  &  $>$24,190  &  $>$28,533  &  \textbf{17,913}  \\
      & Average NV (54) & 298   & 391   & 411   & 250 \\
\hline\hline
\multicolumn{6}{|c|}{\textit{$\ldots$ but {\sc AllS} can still win on such benchmarks}} \\
\multirow{3}{*}{sgb-book} & Completion (26) & 20    & 20    & 20    & 20 \\
      & $\sum$CPU sec. (20) &  9,677  &  \textbf{8,315}  &  8,455  &  8,565  \\
      & Average NV (20) & 143,653 & 148,055 & 148,985 & 134,099 \\
\hline
\multirow{3}{*}{tightness0.1} & Completion (100) & 100   & 100   & 100   & 100\\
      & $\sum$CPU sec. (100) &  46,926  &  \textbf{43,766}  &  44,971  &  69,974  \\
      & Average NV (100) & 10,347 & 9,762 & 9,948 & 12,457\\
\hline
\multirow{3}{*}{ukVg} & Completion (65) & 29   & 31   & 28   & 30\\
      & $\sum$CPU sec. (31) &  $>$19,466  &  \textbf{19,040}  &  $>$20,961  &  $>$19,119  \\
      & Average NV (28) & 141 & 411 & 133 & 139\\
\hline
\end{tabular}
  }
\end{table}
} For these categories of benchmarks the statistical analysis of
Table~\ref{tab:ranking-poac} shows that {\sc Var} performs the best.
Indeed, for full-insertion, tightness0.8, and wordsVg {\sc Var}
has the smallest CPU time of the strategies.
However, individual benchmarks may vary despite the identified
statistical groupings.  For example,
{\sc AllS} performs best on the tightness0.1, sgb-book, and ukVg
benchmark, respectively.

We conclude that, unless we know enough about the problem instance
under consideration, we should use {\sc AllS} in conjunction with
POAC, as the overall analysis shows us.

\begin{figure}[!ht]
\centering
\includegraphics[width=\textwidth]{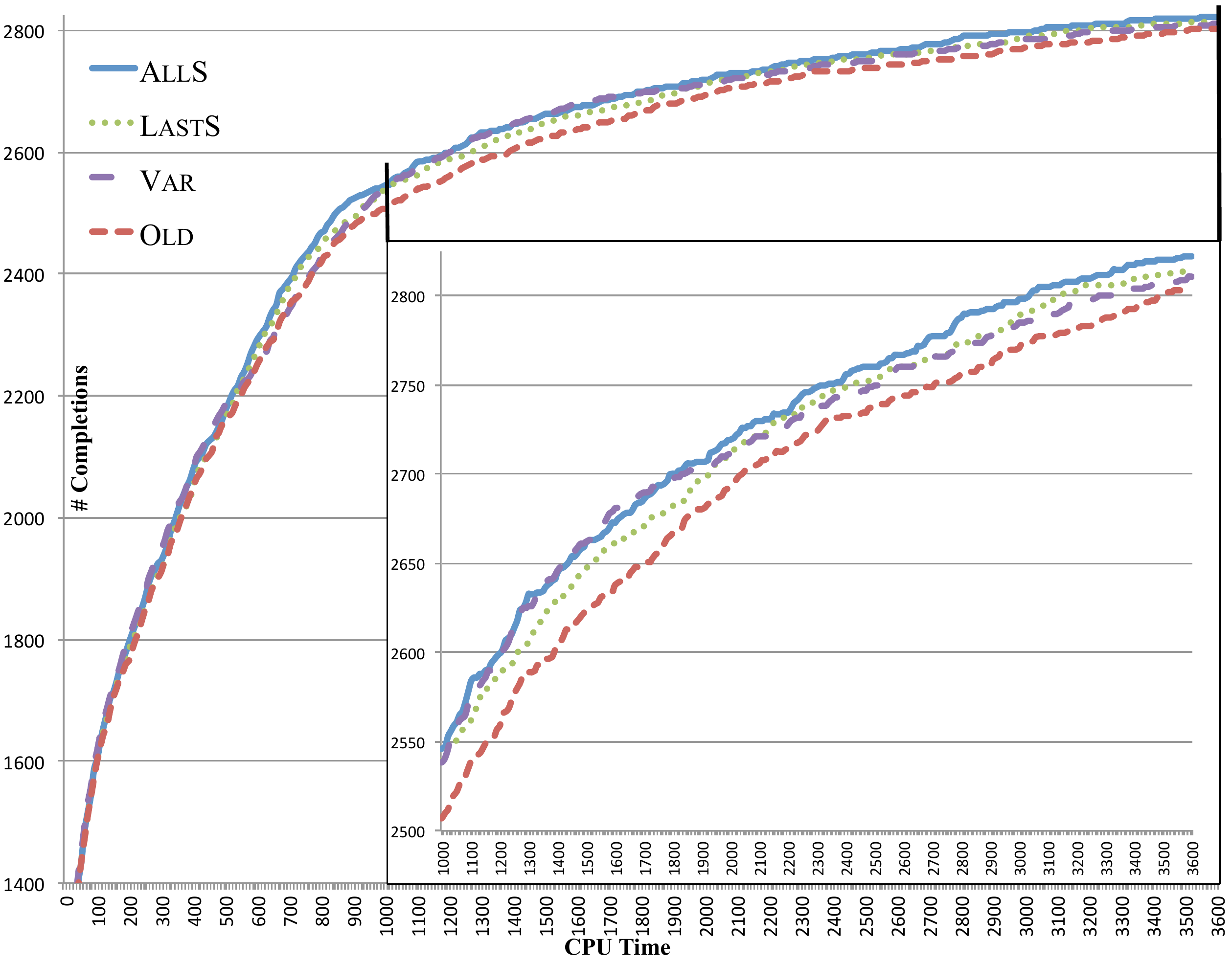}
\caption{\small Cumulative number of instances completed by CPU time
  for POAC}
    \label{fig:poac-time}
\end{figure}

Figure~\ref{fig:poac-time} shows the cumulative number of instances
completed by each strategy as CPU time increases.
For easy instances ($<100$ seconds), the completions of the strategies
are similar.  As the time limit increases {\sc Old} becomes dominated
by the other three strategies.  To better compare {\sc AllS}, {\sc
  LastS}, and {\sc Var} we examine the hard instances, zooming the
chart on the cumulative CPU time solved between 1,000 and 3,600
seconds.  Although {\sc Var} performs well on smaller CPU time ({\sc
  Var} contends with {\sc AllS} for the most completed instances
between 1,000 and 1,700 seconds) it becomes dominated by {\sc AllS}
and {\sc LastS} on the harder instances.  {\sc AllS} clearly dominates
all other strategies.  These curves confirm the results of the
statistical analysis given in Table~\ref{tab:ranking-poac}.

\subsection{Relational Neighborhood Inverse Consistency}
\label{sec:exp-rnic}
The statistical analysis compares the relative performance for {\sc
  Old}, {\sc AllC}, and {\sc Head} for RNIC. It shows that, {\em
  overall}, {\sc AllC} and {\sc Head} are equivalent and {\sc Old}
has the worst performance.  The following holds in general for all
benchmarks:
\begin{equation}
\normalfont\textsc{AllC} \equiv \normalfont\textsc{Head} > \normalfont\textsc{Old}
\label{eq:rnic-ranking}
\end{equation}
\ig{\begin{center}
  \framebox[5cm]{{\sc AllC} $\equiv$ {\sc Head} $>$ {\sc Old}}
\end{center}}
The fact that {\sc Old} is the worst demonstrates that RNIC's
contribution to the weights of dom/wdeg should not be ignored, thus
justifying our investigations.

Table~\ref{table:all-rnic} summarizes the
experiments' results on all the 132 tested benchmarks.
{\sc AllC} is the best strategy on all measures while {\sc Old} is
the worst.
{\def\arraystretch{1.5}
 \begin{table}[!ht]
  \caption{\small Results of experiments for RNIC}
  \label{table:all-rnic}
  \centerline{
    \begin{tabular}{|l|rrr|}\hline
      & \multicolumn{1}{c}{\sc Old} & \multicolumn{1}{c}{\sc AllC} & \multicolumn{1}{c|}{\sc Head} \\
\hline
\hline
\# Completion (3,869) & 2,420 & 2,427 & 2,423 \\
$\sum$CPU sec. (2,416) &  $>$1,032,130  & \textbf{$>$1,010,221} &  $>$1,014,635  \\
Average NV (2,432) & 77,067 & 45,696 & 45,803 \\\hline
\end{tabular}
  }
\end{table}
}

We were not able to uncover meaningful categories of
benchmarks to distinguish between {\sc AllC} and {\sc Head}.
Table~\ref{table:rnic-dimacs} summarizes individual benchmark results
for the Dimacs category.
\begin{table}[!b]
  \caption{\small Examples of Dimacs benchmarks where {\sc AllC}
    and {\sc Head} perform best}
  \label{table:rnic-dimacs}
  \centerline{
\begin{tabular}{|ll|rrr|}\hline
{\bf Benchmark} & \multicolumn{1}{c|}{} & \multicolumn{1}{c}{\sc Old} & \multicolumn{1}{c}{\sc AllC} & \multicolumn{1}{c|}{\sc Head} \\
\hline
\hline
\multirow{3}{*}{pret} & Completion (8) & 4     & 4     & 4 \\
      & $\Sigma$CPU (4) &  196  & \textbf{28} &  61  \\
      & Average NV (4) & 1,285,234 & 125,793 & 273,736 \\
\hline
\multirow{3}{*}{dubois} & Completion (13) & 6     & 9     & 11 \\
      & $\Sigma$CPU (6) &  $>$22,041  &  $>$10,088  & \textbf{1,348} \\
      & Average NV (11) & 11,222,349 & 1,522,902 & 382,329 \\
\hline
\end{tabular}
  }
\end{table}
Within the category, either {\sc AllC} or {\sc Head} perform the best
by all measures on different benchmarks.
Similar results are obtained on the graph coloring category, shown in Table~\ref{table:rnic-gc}.
\begin{table}[!ht]
  \caption{\small Two graph coloring benchmarks where {\sc AllC}
    and {\sc Head} perform best}
  \label{table:rnic-gc}
  \centerline{
\begin{tabular}{|ll|rrr|}\hline
{\bf Benchmark} & \multicolumn{1}{c|}{} & \multicolumn{1}{c}{\sc Old} & \multicolumn{1}{c}{\sc AllC} & \multicolumn{1}{c}{\sc Head} \\
\hline
\hline
\multirow{3}{*}{mug}   & Completion (8) & 8     & 8     & 8 \\
      & $\Sigma$CPU (8) &  5,098  & \textbf{548} &  2,819  \\
      & Average NV (8) & 1,501,379 & 189,595 & 883,130 \\
\hline
\multirow{3}{*}{leighton-15} & Completion (26) & 5     & 5     & 5 \\
      & $\Sigma$CPU (5) &  2,219  &  1,493  & \textbf{1,222} \\
      & Average NV (5) & 25,014 & 12,461 & 4,972 \\
\hline
\end{tabular}
  }
\end{table}
Having such different results between {\sc AllC} and {\sc
  Head} explains why the statistical analysis found them to be
equivalent.  Regardless, either {\sc AllC} or {\sc Head} performs
better than {\sc Old} in a statistically significant manner.

\begin{figure}[!ht]
\centering
\includegraphics[width=\textwidth]{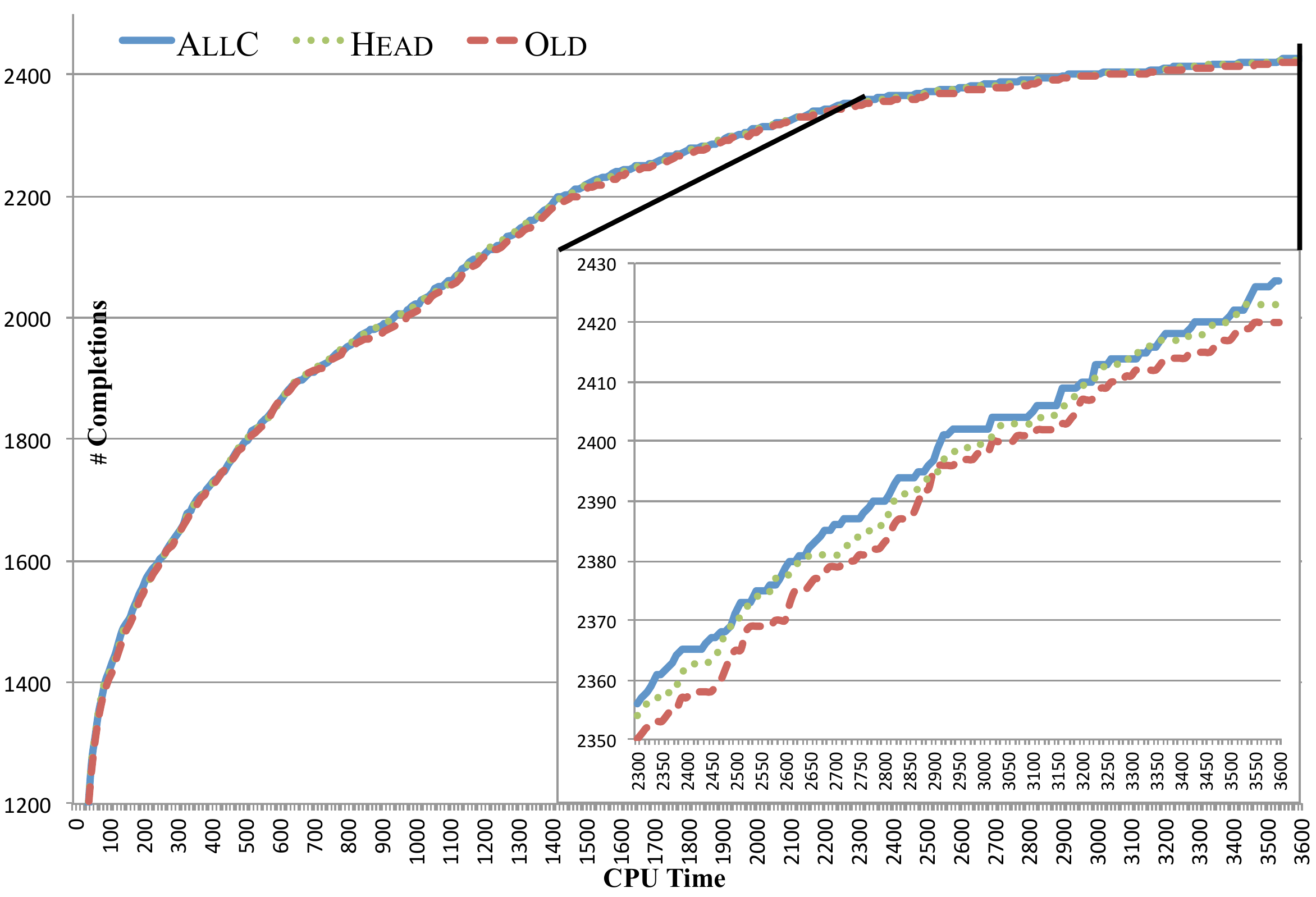}
\caption{\small Cumulative number of instances completed by CPU time
  for RNIC}
    \label{fig:rnic-time}
\end{figure}

Figure~\ref{fig:rnic-time} shows the cumulative number of instances
completed by each strategy as CPU time increases.
As was the case for POAC, on easy instances ($<100$ seconds), the
completions of the strategies are similar.  Focusing on harder
instances, solved between 2,300 and 3,600 seconds, {\sc
  Old} becomes dominated by {\sc AllC} and {\sc Head}.  The curves of
{\sc AllC} and {\sc Head} remain close to one another.  These curves
confirm the ranking in Equation~\ref{eq:rnic-ranking}.

\section{Conclusion}
\label{sec:conclusion}
This paper introduces four strategies for incrementing the weight in
dom/wdeg for singleton consistencies (POAC) and three strategies for
relational consistencies (RNIC).  For both consistencies, {\sc Old} is
the worst strategy and a weighting schema involving the higher-level
consistency is necessary.  We show that for POAC the best method is
{\sc AllS}, which increments the weights at every singleton test. For
RNIC, we show {\sc AllC} and {\sc Head} are statistically equivalent.
Our work is a first step in the right direction, especially given the
importance of higher-level consistencies in solving difficult CSPs.
Future work may need to investigate more complex strategies for these
and other consistencies.

\section*{Acknowledgments}
The idea of \textsc{Var} was proposed by Christian Bessiere.
 This research is supported by NSF Grant
No.\ RI-111795 and RI-1619344.  Experiments were completed utilizing
the Holland Computing Center of the University of Nebraska, which
receives support from the Nebraska Research Initiative.

\bibliography{my-bib-file}{}
\bibliographystyle{named}

\end{document}